\documentclass{article}

\usepackage{arxiv}

\usepackage[utf8]{inputenc} 
\usepackage[T1]{fontenc}    
\usepackage{hyperref}       
\usepackage{url}            
\usepackage{booktabs}       
\usepackage{amsfonts}       
\usepackage{nicefrac}       
\usepackage{microtype}      
\usepackage{cleveref}       
\usepackage{lipsum}         
\usepackage{graphicx}
\usepackage{natbib}
\usepackage{doi}

\usepackage{algorithm}
\usepackage{comment}
\usepackage{algorithm,algpseudocode}

\title{Perturbing the Gradient for alleviating Meta Overfitting}

\author{ 
    Manas Gogoi\\
	Department of Information Technology\\
	Indian Institute of Information Technology\\
	Allahabad \\
	\texttt{pcl2017001@iiita.ac.in} \\
	\And
	Sambhavi Tiwari \\
	Department of Information Technology\\
	Indian Institute of Information Technology\\
	Allahabad \\
	\texttt{rsi2018503@iiita.ac.in} \\
    \And
	Shekhar Verma \\
	Department of Information Technology\\
	Indian Institute of Information Technology\\
	Allahabad \\
	\texttt{sverma@iiita.ac.in} \\
}

\renewcommand{\shorttitle}{\textit{arXiv} Template}

\hypersetup{
pdftitle={Perturbing the Gradient for alleviating Meta Overfitting},
pdfsubject={q-bio.NC, q-bio.QM},
pdfauthor={David S.~Hippocampus, Elias D.~Striatum},
pdfkeywords={Meta Overfitting, Optimization, Meta Learning, MAML},
}

\begin{document}
\maketitle

\begin{abstract}
	The reason for Meta Overfitting can be attributed to two factors: Mutual Non-exclusivity and the Lack of diversity, consequent to which a single global function can fit the support set data of all the meta-training tasks and fail to generalize to new unseen tasks. This issue is evidenced by low error rates on the meta-training tasks, but high error rates on new tasks. However, there can be a number of novel solutions to this problem keeping in mind any of the two objectives to be attained, i.e. to increase diversity in the tasks and to reduce the confidence of the model for some of the tasks. In light of the above, this paper proposes a number of solutions to tackle meta-overfitting on few-shot learning settings, such as few-shot sinusoid regression and few shot classification. Our proposed approaches demonstrate improved generalization performance compared to state-of-the-art baselines for learning in a non-mutually exclusive task setting. Overall, this paper aims to provide insights into tackling overfitting in meta-learning and to advance the field towards more robust and generalizable models.
\end{abstract}


\section{Introduction}

Of the many challenges that exist in the meta-learning setting, one common issue is that of meta-overfitting \cite{huisman2021survey}. Meta overfitting occurs when a single function tries to fit in the examples from all the tasks. This results in memorization of all the task examples and performs poorly when a novel task comes into the picture \cite{yin2019meta}. 

In traditional supervised learning, overfitting occurs as the model captures the inherent noise in the data considering it as a part of the underlying structure. In other words, for any training data, the model tries to learn a complex function that overfits the labels \cite{roelofs2019meta}, \cite{li2023understanding}. Owing to the two interleaved levels of learning in meta-learning, the overfitting corresponding to it is different, as it encompasses overfitting at its different levels and combinations. In other words, there are two types of overfitting involved with meta-learning - memorization and learner overfitting \cite{rajendran2020meta}; the former refers to the case when the model fails to learn to rely upon or utilize the support set and hence performs poorly on the query set, while the latter is the case of standard overfitting that may occur within a particular task, i.e. if the base learner learns a function to fit into all of the support set examples and does not to generalise to the query points of that task.

Although, learner overfitting is more or less equally prevalent in meta learning scenario, due to the nature of the problem and its impact on use cases, its meta-level counterpart is more pronounced and as such this paper deals with meta-overfitting or memorization in the widely discussed MAML framework \cite{finn2017model}.

The analogy of meta overfitting with that of vanilla overfitting is that in traditional supervised learning, overfitting occurs as the model tries to fit into all the data points with a single learned function, and in the case of meta overfitting, the model tries to fit into all the tasks in the meta trainng set with a single function. Thus, while in the ideal scenario the meta learned function is more generic and requires to be fine tuned to the particular task by using the task training data, meta-overfitting attempts to fit all the task specific functions with a single meta learned function without the need for adaptation.



Recent works have tried to solve the problem of meta overfitting. One approach involves modifying the learning model, as presented in \cite{yin2019meta}, while another approach aims to increase task diversity, as described in \cite{rajendran2020meta}, to prevent the model from learning a single global function and consequently eliminating meta overfitting. In all of the existing methods, either the data itself is augmented or the model capacity is reduced. However, our approach is based on the idea that in a non mutually exclusive setting, meta overfitting can be largely eliminated if the model is restrained from being able to learn a single global function or in other words, avoiding the parameters that lead to that scenario. One way of achieving this is by diverting the pathway of the parameters. Our approach which is partly optimization based, involves designing an optimization procedure, specifically for meta overfitting scenarios. It involves optimization as the method requires for the gradients to be altered during optimization.

In this paper, we provide a novel solution to the problem of meta-overfitting by restricting the trajectory of the gradients from moving towards the overfitting parameters. The proposed approach is applicable to all gradient based meta learners, in general. Moreover, in this paper, we introduce a feedback mechanism from the meta-test to the meta-train phase to mitigate the effect of meta-overfitting and provide an emperical study on the same. We compare the results of our proposed method with existing methods for handling meta-overfitting on benchmark datasets.

\section{Problem Definition}
\label{submission}
The key factors that lead to meta-overfitting are Non-mutual exclusiveness and Lack of Diversity. A set of tasks is set to be non-mutually exclusive when the setting of the tasks is such that it is possible for a single function to fit into all the examples in all of the tasks. On the other hand, lack of diversity signifies that the tasks are similar to the extent that they share a common structure and have less variation in their task-specific information. Under the aforementioned conditions of task setting, the model in its course of meta-training, attains a state where the model performs very well on the meta-training set but fails to generalise to any new task from the same distribution. The main reason for this behaviour is that the model implicitly memorizes all the examples from the meta-train tasks rather than ideally learning how to adapt using the support set of a task. Hence, when the model encounters a new task, it simply gives a zero shot prediction for the test input.

\section{Related Work}

The problem of meta overfitting was first defined and discussed in detail in Yin et al \cite{yin2019meta}. According to them, the information content in a memorizing model is larger and more complex than a meta learned model and hence, the issue was alleviated by regularizing the information complexity of the meta learned function. 

Later works such as Tian et al \cite{tian2020meta} have addressed the problem of overfitting in meta learning with pruning the network to have a meta learned subnetwork. This subnetwork has fewer counts of parameters and thereby higher bias than the original network but still performs fast learning with comparable meta-test accuracy.

Moreover, overfitting in the area of meta-learning at large was discussed in Rajendran et al \cite{rajendran2020meta}, while clearly distinguishing between memorization and learner overfitting, the former being the more commonly addressed form. Contrary to the Conditional Entropy (CE) preserving augmentation used to handle standard overfitting, the paper focuses on CE-increasing augmentations that involves the use of a random key governing the input-output relationship function. This encourages the model to make use of the support set, without which it is not possible to learn the same. As compared to Yin et al it does not restrict information to the data, rather targets on increasing task diversity.

Although in approaches such as MetaSVM \cite{lee2019meta}, CAVIA \cite{zintgraf2019fast} etc., the issue of overfitting existing in meta-learning domain is partly handled, it essentially refers to the inner task overfitting rather than meta-overfitting. In MetaSVM, the meta learner encodes the data for the base learner with a convex loss objective to quickly be able to perform the task quickly with the encodings. The base learner generally is a Support Vector Machine and methods such as regularization can effectively mitigate the inner task overfitting. On the other hand, CAVIA seggregates the meta learned parameters from the task specific parameters due to which it does not suffer from overfitting in the inner loop even when the network size is increased. Thus, it is possible to have different expressive powers for both the task specific function and the meta-level function.

Our approach is different from the existing ones as we do not reduce the capacity of the model and neither augment the task data in any way; rather we aim to alter the optimisation process so as to lead the gradient descent trajectory to a different optimum than the one that causes meta-overfitting.

\section{Preliminaries}
\subsection{Meta Learning}
Meta-learning, also referred to as "learning to learn", is a subfield of machine learning that focuses on developing models that can quickly adapt to new tasks \cite{santoro2016meta}, \cite{snell2017prototypical}, \cite{garnelo2018conditional}, \cite{munkhdalai2017meta}, \cite{sung2018learning} . It involves a higher notion of learning with two interleaved levels; at the base level, a model learns to perform any specific task; whereras as the model performs more and more such different tasks, it gains a meta-level knowledge about how to perform any new task in general from that domain and this meta-level knowledge in turn helps to rapidly perform any specific base task from that task distribution. Given a set of $T$ tasks, each with a corresponding dataset $D_i$, a meta-learning algorithm aims to learn a function $f(D_i, \theta)$, where $\theta$ are the parameters of the model, that generalizes well to new tasks sampled from the same distribution.

One approach to meta-learning is to learn an initialization of the parameters $\theta$ that can be quickly fine-tuned on new tasks with a small number of gradient updates. This approach, known as model-agnostic meta-learning (MAML) \cite{finn2017model}, involves optimizing the following objective function :

$$ \min_{\theta} \sum_{i=1}^{T} \mathcal{L}{i}(f(D{i}, \theta_{i}^{\prime})) $$

where $\mathcal{L}{i}$ is the task-specific loss function and $\theta{i}^{\prime} = \theta - \alpha \nabla_{\theta} \mathcal{L}{i}(f(D{i}, \theta))$ is the updated parameter set after a few gradient updates on task $i$.

Meta Learning has applications in various domains such as few-shot image segmentation \cite{luo2022meta}, \cite{hendryx2019meta}, robotics \cite{song2020rapidly}, \cite{gupta2018meta}, \cite{nagabandi2018learning}, Natural Language Processing \cite{zhou2021bert}, \cite{yu2018diverse}, \cite{ye2019multi} recommender systems \cite{wang2021preference}, \cite{chen2021improving}, \cite{cunha2016selecting} etc

\subsection{Non-Mutual Exclusivity}
Non mutual exclusivity in meta-learning refers to a property of the task distribution such that it is possible for a single function to fit across all the examples in all of the tasks \cite{yin2019meta}. Ideally, this may seem plausible but optimizing for the meta-level objective under the scenario, drives the meta learning model parameters to a point where the loss may be minimum but relies wholly on memorization of the entire meta training data set. This prevents the model from learning generalizable knowledge and shared structure across the tasks. In other words, the model fails to utilize the support set for accommodating task specific knowledge and gives zero shot prediction for the task query set. Thus, the model performs very well during meta training since it overfits all the meta training support and query set data in the learnt function of high complexity, but fails drastically when it comes to learning a new task at meta test. This defeats the basic idea of meta learning i.e. to develop a learning model that can perform an unseen task quickly.

Examples of non-mutual exclusive task settings include the problem of sinusoidal regression under the given conditions : each task is a sinusoidal regression task with an amplitude A and phase difference $\phi$ sampled from some fixed distribution but under the condition that the domains for each task are disjoint from one another with an equally sized gap in between. Because of the gap between the domains, the entire setting becomes non-mutually exclusive as it permits a single continuous function to be learnt that perfectly maps the data points in the domain intervals corresponding to a task and learns an irrelevant mapping in the gaps.

Another example of non-mutual exclusivity is the task of pose prediction. Given a few images (x) of a rotated object and the relative degree of rotation (y), the objective is to determine the degree of rotation for a new image of the object. The premise of meta learning arises from the accumulation of knowledge with performing such predictions for a multitude of different objects. However, this task arrangement is considered to be non-mutually exclusive and is attributed to the fact that the few count and distinctive feature of the different objects in the dataset makes it easy for the model to memorize the default orientations of each of them. Hence it fails to perform when a new object is shown for prediction.

For the case of classification, a non-mutually exclusive task can be designed as if the partitioning of tasks is done in an ordered manner to maintain consistency of the class index throughout epochs i.e., for any particular class the corresponding label is globally same across tasks, whenever it appears. This is non-mutually exclusive because, the model can now potentially memorize the global class-label binding for the classes that appear continually in the meta-training phase and quite obviously fail to generalize to tasks with unseen classes.

\section{Proposed Methodology}
In optimization based meta-learning \cite{nichol2018reptile}, \cite{rajeswaran2019meta}, \cite{grant2018recasting}, \cite{finn2019online}, \cite{rusu2018meta} the model optimises its objective function over two loops, usually involving Hessian computations and gradually moves towards the parameters with minimum loss. Once converged, these parameters are optimal for all the tasks in the task distribution, considering that only a few steps of fine tuning with the support set gives the task specific parameters for any task. Hence, any previously unseen task from the distribution can be effectively solved with minimal adaptation. However, in non-mutually exclusive settings, optimising the model to reduce the loss results in a memorized function that meta overfits to the tasks used to train the model and does not generalize to unseen tasks. This is due the non-mutually exclusive nature of tasks used for training that causes the model to learn a single function to fit all training tasks. In other words, the optimizer naturally leads the trajectory of the meta-parameters towards minimum loss and ends up at the meta-overfitting parameters that give minimum training task loss but significantly higher test task loss. Our methodology is based on preventing the consequences of non-mutual exclusivity rather than preventing non-mutual exclusivity altogether. The idea is to prevent the model parameters from reaching the minima of the loss hyperspace or in other words, prevent the model to learn a single parameterised function that best fits all the meta training tasks. The loss surface in terms of the model parameters being non-convex, has more than one local optima and as such our objective is to find a different set of optimal parameters than the one suggested plainly by the optimizer. In our case, this is ensured by augmenting the optimisation method.

The proposed optimisation method involves adding a noise term to the gradient direction at each step of optimisation. Due to its statistical properties and general usefullness, the noise is considered to be drawn from a Gaussian with a zero mean and non-trivial standard deviation. This addition of the noise term to the gradient can also be considered as incrementing the diversity of the tasks drawn from the distribution, as the gradient direction of each task is changed from its default. This diverts the model parameters from moving in a deterministic direction towards the minima given by the gradient descent process. As a result, due to perturbations of the gradient at any time step, the gradient at the next step gets altered from its otherwise supposed direction and eventually all subsequent steps are influenced. Thus, in our methodology, the effective gradient becomes probabilistic in nature and is governed by the resultant noise accumulated over all of the steps. However, to enforce convergence of the model the noise is scaled down by a factor after a certain number of iterations. This optimisation method can be employed in both the inner loop or the outer loop of MAML as well as simultaneously. 

\[ \hat{\phi^{'}_{i}}=\phi_{i} - \nabla_{\phi}Loss(D^{tr})  + \epsilon \]
where \[\epsilon \sim \mathcal{N}(\mu, \sigma) \]

To explain the working of this approach, we can assume a 3d landscape with hills and valleys that represent the meta-loss landscape over a 2d parameter space, such that a valley at its lowest point represent the minimum loss in some proximity and the corresponding co-ordinate of its location represents a locally optimal parameter. Starting from any random point in the landscape and traversing along the direction of gradient, tends to move towards the a nearby valley i.e. a point where the loss is minimum. Ideally, this parameter is considered optimal for the task at hand. However, under the condition of non-mutually exclusive task-setting, location of the deepest valley or point of minimum loss is by design, the parameters that encourages memorization. Thus, under such circumstances, the model suffers from overfitting. On the other hand, when we add a Gaussian noise to the direction of gradient descent, it diverts the trajectory from its pre-determined path and causes it to probabilistically arrive at some other valley, the parameters for which may not be the most optimal in terms of training loss but which will be sufficiently optimal to perform well on the testing tasks and at the same time prevent the model from overfitting. This in turn enables generalization to new tasks because in this case relying on a single global function does not minimize loss for the model and is hence encouraged to utilize the support set.

From the perspective of model objective, meta-overfitting occurs as the task setting allows for a single universal function to fit all of the tasks in due course of loss minimization. This loss minimization happens iteratively by following the gradient prescribed by the optimizer and consequently the model tends to learn the universal function. However, in our method, as we perturb the gradient with noise, the model is unable to attain minimal loss by learning a single function that memorizes all the tasks. This in turn prevents the model from meta-overfitting and the model has to rely on the support set of a task for fine-tuning the meta-level parameters and finding the task specific function that guarantees minimum loss.

Another way of viewing the mechanics of the methodology is from the perspective of adding task diversity. The difference between traditional machine learning and meta learning is that in machine learning we aim to maximize the log probability of the parameters given the training datapoints. However, in meta learning, we maximize the log probability of the meta-parameters given the meta-training datasets or tasks. In other words, an important observation here is that tasks in meta learning are analogous to datapoints in machine learning. In machine learning, overfitting can be mitigated in classification tasks by increasing the entropy of the input data given a label. The datapoints are augmented to generate more number samples due to which the diversity in the dataset increases. For image related tasks, this augmentation can be in the form of rotations, cropping or flipping of images with certain randomness, which can be considered a function governed probabilistically by a noise. By that same analogy, meta-overfitting in meta-learning can be mitigated by augmenting the tasks or adding diversity in the set of tasks. One form of task augmentation can be the change in the position of optimal parameters for a particular task. In our approach, instead of directly augmenting tasks to have a changed position in the parameter space, we augment the optimization trajectory that inherently leads to the changed parameter. 



\begin{figure}[h]
     \centering
     \includegraphics[width=0.5\textwidth, clip, keepaspectratio]{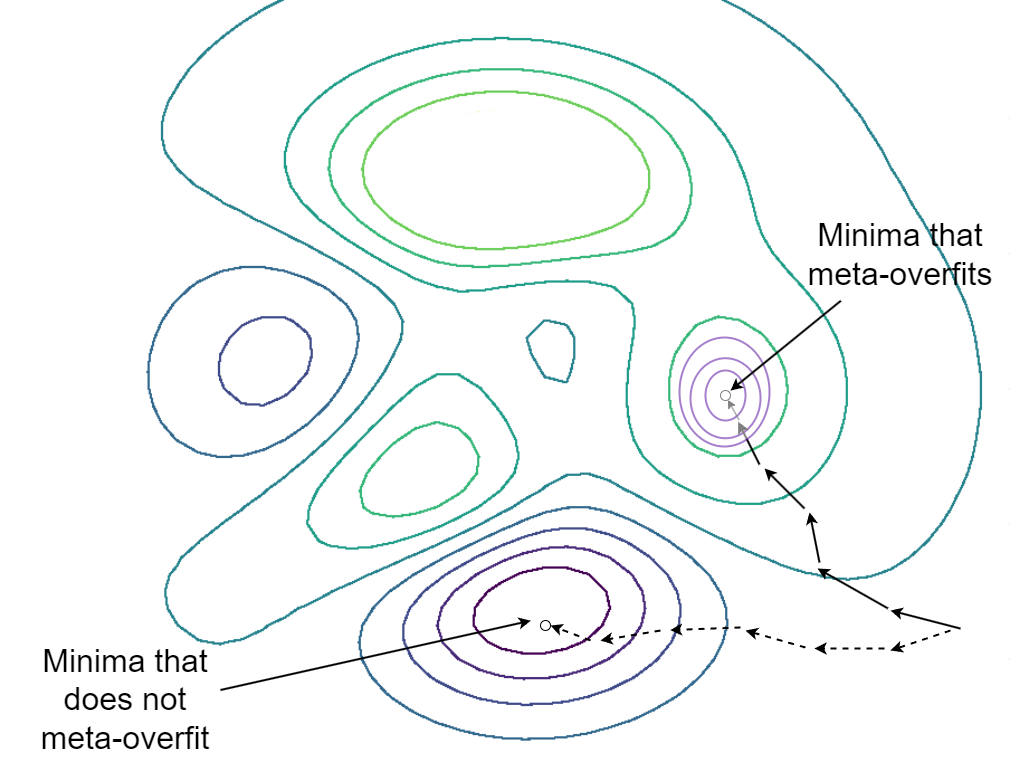}
     \caption{Schematic diagram of the proposed approach. Under the condition of non-mutually exclusive tasks, the optimization process (shown in solid lines) leads the trajectory to the minima in space that meta-overfits or memorizes. However, with the addition of noise, as in our approach, the optimization trajectory (shown in dotted lines) probabilistically arrives at some other point of minima that does not meta-overfit}
     \label{fig:parameter_space}
\end{figure}

\begin{algorithm}[h]
   \caption{MAML Algorithm for Meta Overfit}
   \label{alg:algo}
\begin{algorithmic}[1]
   \State {\bfseries Input:} Meta-training tasks $T_{m}$ sampled from task distribution $p(D^{tr})$, inner loop learning rate ($\alpha$), outer loop learning rate ($\beta$)
   \State {\bfseries Output:} optimal parameters $\theta^{*}$.
   \State Initialize the network $f$ with random $\theta$.
   \State Sample mini-batch of tasks $\{T_{i}\}^{s}_{i=1}$ from meta training tasks $T_{m}$.
   \For{ each task  $T_{i}$ }
    \State Evaluate $\nabla_{\theta}\mathcal{L}_{D^{sup}_{T_{i}}}(f_{\theta})$
    \State Compute adapted parameters using Augmented SGD: \hspace*{6mm} $\phi_{i}$ = ${\theta - \alpha \nabla_\theta \mathcal{L}_{D^{sup}_{T_{i}}}(f_{\theta})} + \epsilon$, \hspace*{1mm} where $\epsilon \sim \mathcal{N}(\mu, \sigma)$ 
   \EndFor
   \State  Update $\theta^{'} \leftarrow \theta - \beta\sum_{T_{{i}\sim p(T)}}\nabla_{\theta_{i}} \mathcal{L}_{T_{i}}(f_{\phi_{i}}) $
\end{algorithmic}
\end{algorithm}

\section{Feedback from Meta-Test to Meta-Train}
Meta overfitting can also be viewed and indicated from the inability of a meta learning model to perform well on an unseen task during meta-testing but comparatively much better performance on meta-training tasks. Thus, to close the disparity between the two scenarios, we propose to change the evaluation setting and allow the newly seen task to provide a feedback to the meta-training phase. This feedback mechanism acts as crucial information for the model to adapt the meta-parameters during re-meta-training so as to escape from the valley of the overfitting parameter instance. This allows for an improved performance on that particular unseen task. This is substantial because whenever the model overfits, it fails to fine-tune to any task from the meta-test set which comprises of classes that are mutually exclusive or completely unseen in the meta-train. Fine-tuning extensively further leads to overfitting to the few samples within the specific task. However, the feedback mechanism proposed here allows a model to recover and perform well on any desired task from the meta-test after fine-tuning, while maintaining generality of the samples within the task.

In the proposed feedback mechanism, the parameters initially obtained after meta-training process are further retrained to move in the direction of the optimal test task parameters. For this purpose, the model is tested on a meta-test task to evaluate its performance which tends to be suboptimal in the meta-overfitting case and simultaneously the gradient of the test task loss at the meta learnt parameters is recorded. This indicates the direction in which the meta-parameters must have essentially been for performing well on the test task. We hence leverage on this gradient direction to move the meta-parameters towards the test-task optimal parameters by virtue of retraining the meta-trained model with more meta-training tasks. For every task in the meta-training set, the outer loop gradient is weighed according to its cosine similarity with the test task gradient. Hence, the meta-training tasks that exhibit gradients similar to the test task will be given higher weightage and thereby pulling the updated meta-parameters towards the gradient direction given by the test task. This proposed method enables the model to climb out of the overfitting parameters and move towards a point in the parameter space which is more suited for the test task. Moreover, this approach is also space efficient as there is no need for storing the test task throughout the retraining and merely the gradients need to be stored for comparison. The algorithm is given in \ref{alg:algo1}. Hence by using the feedback mechanism, performance on a particular task can be improved even in the case of meta-overfitting.

\begin{algorithm}[h]
   \caption{Feedback Algo from Meta-Test to Meta-Train}
   \label{alg:algo1}
\begin{algorithmic}[1]
   \State {\bfseries Input:} Meta-training tasks $T_{m}$ sampled from task distribution $p(D^{tr})$, inner loop learning rate ($\alpha$), outer loop learning rate ($\beta$), meta-trained optimal parameters $\theta^{*}$.
   \State {\bfseries Output:} Test-task specific meta-parameters $\theta^{*}$.
   \State Initialize the network $f$ with meta-trained parameters $\theta^{*}$.
   \State Sample a new task $T_{new}$ from meta-test set 
   \State Compute gradient for the new task \par
   \hskip\algorithmicindent  $\nabla_\theta \mathcal{L}_{D^{tr}_{T_{new}}}(f_{\theta})$
   \While {not done}
   \State Sample mini-batch of tasks $\{T_{i}\}^{s}_{i=1}$ from meta \par training tasks $T_{m}$.
   \For{each task  $T_{i}$}
    \State Evaluate $\nabla_{\theta}\mathcal{L}_{D^{tr}_{T_{i}}}(f_{\theta})$
    \State Compute inner parameters using SGD: \par
    \hskip\algorithmicindent $\phi_{i}$ = ${\theta - \alpha \nabla_\theta \mathcal{L}_{D^{tr}_{T_{i}}}(f_{\theta})}$
    \State Compute outer gradients using query set \par
    \hskip\algorithmicindent $grad_i = \nabla_\theta \mathcal{L}_{D^{ts}_{T_{i}}}(f_{\phi_i})$
    \State Compute similarity with test task gradient \par
    \hskip\algorithmicindent $h_i = cos\_sim(grad_i, test\_grad)$
   \EndFor
   \State  Update $\theta^{'} \leftarrow \theta - \beta\sum_{T_{{i}\sim p(T)}}h_i * \nabla_{\theta_{i}} \mathcal{L}_{T_{i}}(f_{\phi_{i}}) $
   \EndWhile
\end{algorithmic}
\end{algorithm}

\section{Experiments and Results}
The experiments are performed on an Intel(R) Xeon(R) CPU E5-2630 v3 processor running at 2.40GHz with 16GB RAM and Quadro K2200 Graphics controller. Throughout the experiments, we aim to address that a model can be prevented or recovered from meta-overfitting with the methods described in this paper. We examine the results of our proposed approach and make comparisons with existing methods that deal with meta-overfitting, on both regression and classification tasks. For regression problems, we consider sinusoidal regression and pose prediction tasks whereas for classification we consider the Omniglot, MiniImageNet and D'Claw datasets under a non-mutually exclusive task setting. For formulation of non-mutually exclusive settings, this paper draws inspiration from the methodology described in the experiments of \cite{rajendran2020meta}.

\subsection{Model Architecture}
The base architecture of the neural network model used for both classification and regression is same as the implementation of vanilla MAML by \cite{finn2017model}. The choice for MAML as the base architecture is due to its extensive study in the research community and widespread applicability. The classification model used in the experiments consists of 4 convolutional blocks followed by a fully connected classifier layer. Each convolutional block comprises of a conv2d layer with 64 filters of size 3x3, a batch normalization layer, a max pooling layer and a ReLU activation. The neural network model used for classification is composed of 2 hidden layers with 40 neurons along with ReLU used for nonlinearities. For all the experiments, the initialization parameters for meta-training are drawn from a truncated normal distribution with 0 mean and 0.01 standard deviation.

For perturbing the gradients, we augment the optimization procedure. This is implemented in a straightforward manner by adding a noise term to the usual gradient computed through automatic differentiation in the model with respect to the loss. The noise term is sampled from a normal distribution and is a hyperparmater specific to a dataset. 

In the Feedback method experiments, initially the direction of the gradients for a test task is determined and stored. Now while retraining the model, all the training tasks are weighed according to the cosine similarity of its gradient and the stored direction. This gradually pulls the model parameters in the direction of the test task. The code for the experiments is available at \href{https://github.com/manasgg44/meta_overfit}{Github}.

\subsection{Sinusoid Regression}
For sinusoid regression tasks, we consider a sine wave in one dimension given by $y=Asin(x-\phi)$, where $A$ is the amplitude and $\phi$ is the phase difference. Both $A$ and $\phi$ are sampled from a uniform distribution given by $A \sim [0.1, 5.0]$ and $\phi \sim [0, \pi]$. The domain is restricted to [-5, 5] and is divided into disjoint intervals [-5, -4.5], [-4, -3.5], ....[4, 4.5] for creating non-mutually exclusive tasks. Each interval is assigned a different task with a different pair of ($A; \phi$). For each task, the inputs x are sampled from one of the disjoint intervals and the label y is generated based on the sine wave for the interval.

MAML on the above non-mutually exclusive setting gives poor results but on the other hand, our method in this case significantly recovers from meta-overfitting. For this, the optimizer is augmented with a noise sampled from a normal distribution with zero mean and having a standard deviation of $7\times10^{-7}$ and $2\times10^{-7}$ respectively for the inner and outer loops. Table \ref{table-sine} shows that the MSE can be considerably improved with our proposed approach as compared to plainly using MAML.

\begin{figure}[h]
     \centering
     \includegraphics[width=0.8\textwidth, clip, keepaspectratio]{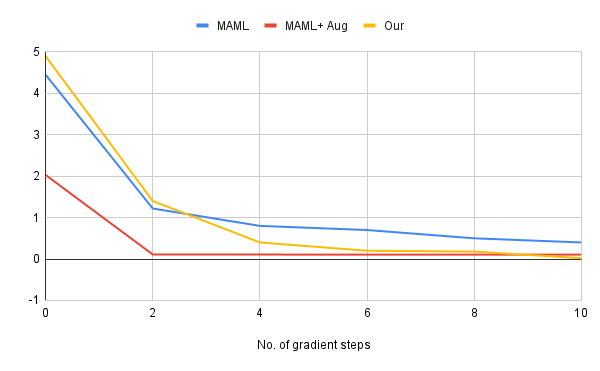}
     \caption{MSE Loss for Sinusoid regression tasks under the non-mutually exclusive task setting averaged over 10 trials.}
     \label{fig:sinusoid}
\end{figure}

In the figure \ref{fig:sinusoid}, it can be seen that with the addition of noise to the optimization, the MAML method starts with a high MSE loss for a small gradient step compared to the baseline and state of the art approach. However, the loss significantly improves as the number of gradient steps is increased. The MSE loss for 1 step and 2 steps is observed to be noisy often compared to the 8, 9 and 10 steps. This shows that proposed approach can recover from a meta-overfitting scenario with more steps of descent.

\begin{table*}[]
\caption{MSE for different approaches on sinusoidal regression task averaged over 10 trials}
\label{table-sine}
\vskip 0.15in

\begin{center}
\begin{small}
\begin{sc}
\begin{tabular}{l|c|c}
\toprule
Approach  & 5 shot  \\
\midrule
MAML    & 0.15  \\
MAML (Meta augmentation) & 0.04  \\
Our     & 0.02 \\
Feedback (Our) & 0.02 \\
\bottomrule
\end{tabular}
\end{sc}
\end{small}
\end{center}
\vskip -0.1in
\end{table*}

\subsection{Classification on Omniglot, MiniImageNet and D'Claw }
For classification, the task formulation is done in two different ways [Meta-Augmentation] based on the level of randomness or the conditional entropy : Intershuffle and Non-mutually exclusive. The intershuffle mode is the default way of task formation in few shot learning, wherein for k way classification, the sampled k classes in any task are randomly assigned a discrete label from [0, k-1] and a class may have different labels when occurring in different tasks. In case of non-mutually exclusive, the classes are divided into k-sized disjoint partitions, and while creating a task from a particular partition, labels are assigned to the task in an ordered fashion. This ascertains that any specific class is always assigned the same label. As a result, the model gets to learn the class and label binding which it memorizes to efficiently generate lower loss during training.  

For our experiments on the few shot classification problem, we consider three benchmark datasets Omniglot, MiniImageNet and D'Claw. Further we evaluate MAML, TAML, MR-MAML along with our approach on the above datasets under the same task settings and provide comparisons. The evaluation metric for all the classification tasks is considered to be accuracy, as the support and query set for all the tasks are designed to be balanced in terms of class label.

The Omniglot dataset [\cite{lake2019omniglot}] comprises of 50,000 handwritten characters as 105x105 grayscale images from 50 different alphabets, including modern and ancient scripts. Multiple individuals wrote each character, producing variations in writing style, orientation, and stroke thickness. The dataset is partitioned into a "background" set of 30 alphabets with 964 characters each and an "evaluation" set of 20 alphabets with 659 characters each. No character appears in both sets, ensuring there is no overlap between the two. The Omniglot dataset is augmented with rotations by multiples of 90 degrees, as proposed by \cite{santoro2016meta}.

For the Omniglot dataset, the tables \ref{table-omni} and \ref{table-omni1} shows the results of classification in the non mutually exclusive case. For both the 20 way and 5 way setting, as compared to vanilla MAML and TAML, our proposed approach is able to improve the classification accuracy significantly and gives comparable results with MR-MAML. For both the cases, the standard deviation of the noise is set to a scale of $10^{-7}$. In comparison to vanilla MAML, where the model gets stuck in a overfitting minima preventing generalisation, perturbing the gradients using our approach allows the model to lift itself from the meta-overfitting minima and converge to another minima in the parameter space.

\begin{table*}[]
\caption{Classification accuracies for different approaches on non-mutually exclusive setting of Omniglot dataset averaged over 10 trials}
\label{table-omni}
\vskip 0.15in
\begin{center}
\begin{small}
\begin{sc}
\begin{tabular}{l|c|c}
\toprule
Approach  & 20way 1shot & 20way 5shot \\
\midrule
MAML    & 7.8   & 50.7  \\
TAML    & 9.6   & 67.9  \\
MR-MAML & 83.3  & 94.1  \\
Our (inner loop)     & 82.36 & 94.2   \\
Our (outer loop)     & 82.36 & 94.1   \\
Our (both loops)    & 82.36 & 94.2   \\
Feedback (ours) & 82.34 & 94.1 \\
\bottomrule
\end{tabular}
\end{sc}
\end{small}
\end{center}
\vskip -0.1in
\end{table*}

\begin{table*}[]
\caption{Classification accuracies of our approach on non-mutually exclusive setting of Omniglot dataset averaged over 10 trials}
\label{table-omni1}
\vskip 0.15in
\begin{center}
\begin{small}
\begin{sc}
\begin{tabular}{l|c|c}
\toprule
Approach  & 5way 1shot & 5way 5shot \\
\midrule
MAML   & 98.1   & 98.7  \\
Our (inner loop)  & 98.3   & 98.8  \\
Our (outer loop)  & 98.3   & 98.8  \\
Our (Both loops)  & 98.5   & 98.9  \\
Feedback (our) & 98.6 & 98.9 \\
\bottomrule
\end{tabular}
\end{sc}
\end{small}
\end{center}
\vskip -0.1in
\end{table*}

The Mini-ImageNet dataset [\cite{vinyals2016matching}] is a subset of the ImageNet dataset with 100 classes, 600 images per class, and with each image being an RGB image of 84x84 pixels. The dataset depicts everyday objects, animals, and natural scenes and is divided into 64 classes for training, 16 for validation, and 20 for testing purposes.

For the case of MiniImagenet dataset, 20 way classification tasks are considered and the results are reported in table \ref{table-mini}. Although the MiniImagenet dataset is quite more complex than the Omniglot, our proposed approach is able to increase the classification accuracy from its vanilla counterpart. Also, using the Feedback algorithm, our model is able to alleviate from the memorization parameters and is able to improve performance on the test task.

\begin{table*}[]
\caption{Classification accuracies for different approaches on non-mutually exclusive setting of MiniImagenet dataset}
\label{table-mini}
\vskip 0.15in
\begin{center}
\begin{small}
\begin{sc}
\begin{tabular}{l|c|c}
\toprule
Approach  & 5way 1shot & 5way 5shot \\
\midrule
MAML    & 26.3   &  41.6  \\
TAML    & 26.1   & 44.2  \\
MR-MAML & 43.6  & 53.8  \\
Our (inner loop)     & 32.36 & 48.25  \\
Our (outer loop)     & 31.89 & 48.02  \\
Our (both loops)     & 32.59 & 48.67  \\
Feedback (our)       & 34.83 & 49.27 \\
\bottomrule
\end{tabular}
\end{sc}
\end{small}
\end{center}
\vskip -0.1in
\end{table*}

The D'Claw dataset is an image classification dataset with 20 different classes, where each class has 35 images. The objective is to classify if an object placed between the fingers of the claw is in the correct orientation or not. It is designed for 2-way classification (0 for incorrect, 1 for correct). The dataset is inspired by robotics and uses hardware designs from ROBEL[\cite{ahn2020robel}].

In case of the D'Claw dataset, the classification accuracy using our approach increases from 72.5 to 73.1 percent in the standard 1 shot setting.

\begin{table*}[]
\caption{Classification accuracies of our approach on non-mutually exclusive setting of Dclaw dataset}
\label{table-dclaw}
\vskip 0.15in
\begin{center}
\begin{small}
\begin{sc}
\begin{tabular}{l|c|c}
\toprule
Approach  & 2way 1shot \\
\midrule
MAML    & 72.5  \\
Our (inner loop)     & 82.36   \\
Our (outer loop)     & 82.36   \\
Our (both loops)    & 82.36  \\
Our     & 73.1  \\
\bottomrule
\end{tabular}
\end{sc}
\end{small}
\end{center}
\vskip -0.1in
\end{table*}

\subsection{Effect of noise addition on the accuracy :}
\begin{figure}[h]
     \centering
     \includegraphics[width=\textwidth, clip, keepaspectratio]{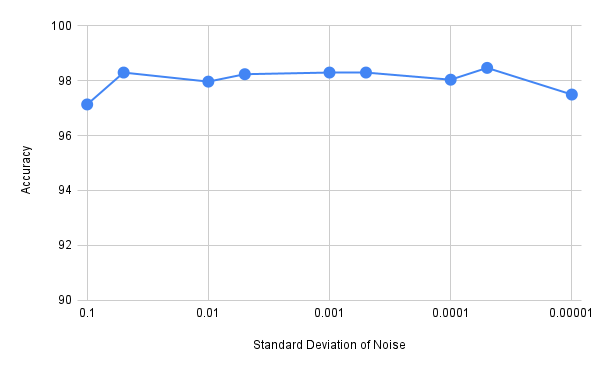}
     \caption{Variation of accuracy for 5 way 1 shot Omniglot dataset classification with varying standard deviation of the noise distribution}
     \label{fig:omni_noise}
\end{figure}

To check the effect of variation in noise on the accuracy of the model, we consider the Omniglot dataset under the 5way 1 shot setting and repeat the experiment with a different noise distribution for each run. At every instance of the experiment, the standard deviation of the noise is varied keeping the mean at 0, and it iterates from $10^{-1}$ to $10^{-5}$ as per a logarithmic scale. Also another series with a logarithmic scale, from $5$x$10^{-2}$ to $5$x$10^{-5}$ is used to accomodate the values in between the former one. For each of the chosen noise distributions, 10 iterations are run with the noise added in the inner loop and the average classification accuracy is reported. Figure \ref{fig:omni_noise} shows the variation of classification accuracy against the standard deviation of the noise distribution. It can be seen that the accuracy of the model somewhat plateaus in the mid range of values and has the highest accuracy in one particular portion of this region. However, the model has lower accuracies towards both the extreme ends of values for the noise axis. Finding the optimal value for the standard deviation of the noise is same as a hyperparameter search and involves a bruteforce method, although patterns can be recognised for the approximate range of values with observation.

\section{Conclusion}
This work is based on the assumption that in order to alleviate from meta-overfitting, the model should be prevented from following its conventional gradient direction in the parameter space that leads it towards a point of minimum loss. In light of the above, we proposed an approach that augments the optimization process to add a subtle noise from a noise distribution, in the gradient direction proposed by the vanilla optimizer. Experiments have shown that the proposed model is able to handle non-mutually exclusive task settings and performs better than the existing state-of-the-art methods for meta-overfitting. The benefit of our approach is that the method does not make any changes to the learning capacity of the model by regularization nor to the data, but only to the optimization procedure to reach at a different optimum altogether. The standard deviation of the noise distribution is observed to be a hyperparameter, a careful selection of the values of which, results in optimized performance and shows a diminishing performance towards either ends of the scale of values. Additionally, we have also introduced a mechanism for providing feedback from the meta-test to the meta-train phase in order to ensure improved performance on a test task drawn from the meta-testing classes under a meta-overfitting scenario. The feedback guides the model parameters in the direction of the test task to a minima that is closer to the specific task. This improves the performance of the model and alleviates meta-overfitting.

\bibliographystyle{plainnat}
\bibliography{references}  






\end{document}